\documentclass{svproc}
\pdfoutput=1

\usepackage{graphicx}
\usepackage{subfig}
\usepackage{caption}

\usepackage{amsmath} % assumes amsmath package installed
\usepackage{amssymb}  % assumes amsmath package installed

\begin{document}
\mainmatter              % start of a contribution
\title{Feature Based Potential Field for Low-level Active Visual Navigation}
\titlerunning{Feature Based Potential Field for Low-level Active Visual Navigation}
\author{R\^{o}mulo T. Rodrigues\inst{1} \and  Meysam Basiri\inst{1} \and A. Pedro Aguiar\inst{2} \and Pedro Miraldo\inst{1} }

\authorrunning{R\^{o}mulo T. Rodrigues et al.} % abbreviated author list (for running head)
\tocauthor{R\^{o}mulo T. Rodrigues$^{1}$,  Meysam Basiri$^{1}$, A. Pedro Aguiar, and Pedro Miraldo}
\institute{Institute for Systems and Robotics, Lisbon, Portugal,\\
\email{\{romulo.rodrigues, meysam.basiri, pedro.miraldo\}@tecnico.ulisboa.pt},
\and
Faculty of Engineering University of Porto, Porto, Portugal, \\
\email{pedro.aguiar@fe.up.pt},
}

\maketitle              % typeset the title of the contribution

\begin{abstract}
This paper proposes a novel solution for improving visual localization in an active fashion. The solution, based on artificial potential field, associates each feature in the current image frame with an attractive or neutral potential energy. The resultant action drives the vehicle towards the goal, while still favouring feature rich areas. Experimental results with a mini quadrotor equipped with a downward looking camera assess the performance of the proposed method.
\end{abstract}
%
%%%%%%%%%%%%%%%%%%%%%%%%%%%%%%%%%%%%%%%%%%%%%%%%%%%%%%%%%%%%%%%%%%%%%%%%%%%%%%%%
\section{Introduction}
To enable the deployment of autonomous Micro Aerial Vehicles (MAVs) in GPS-denied environments or in scenarios where external systems such as motion tracking cameras are not available, the localization problem, that is, the problem of estimating the pose of the MAV with respect to its environment, needs to be adequately addressed. Research community has been focused in using lightweight onboard sensors. Most effective solutions combine Visual Odometry (VO) algorithms and inertial data within a filtering framework \cite{loianno16,weiss13,engel12}. 

\begin{figure}[t!]
  \centering
  \subfloat[]{\includegraphics[width=0.49\linewidth]{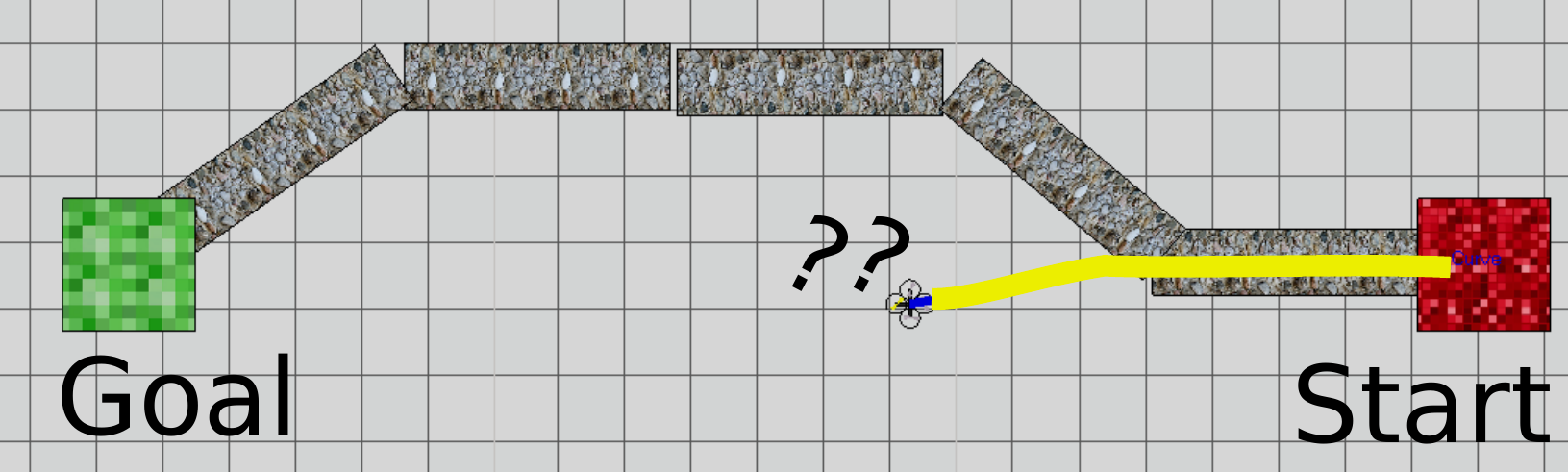} \label{fig:slam_fails}}
  \subfloat[]{\includegraphics[width=0.49\linewidth]{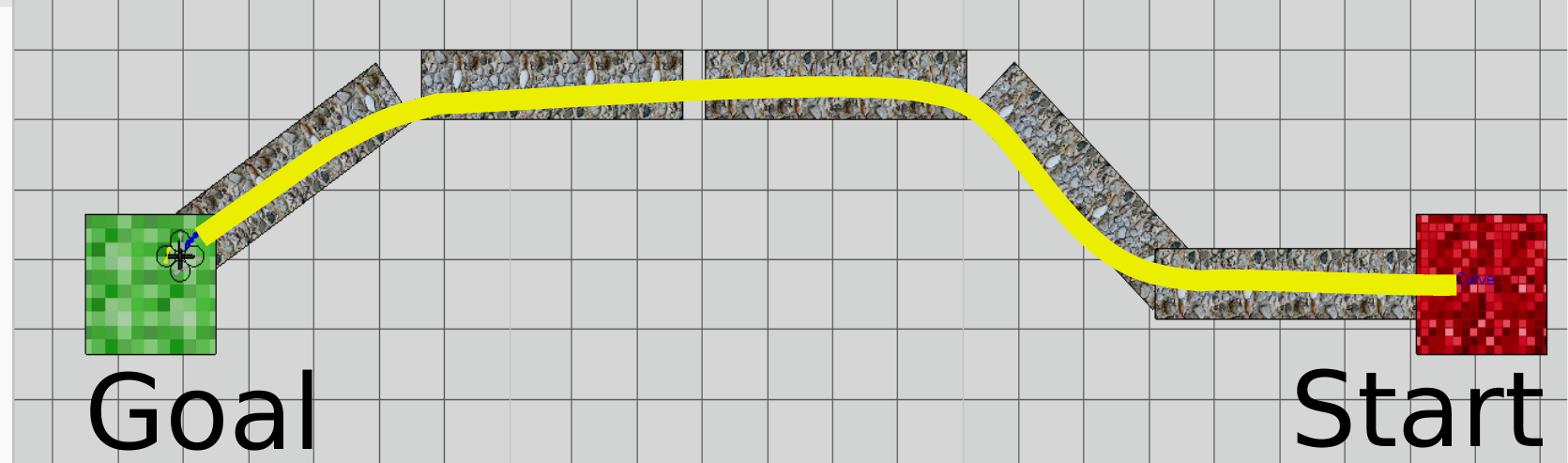} \label{fig:slam_works}}
    \caption{Localization may fail when not considering the perception pipeline in the motion planning and control frameworks \protect\subref{fig:slam_fails}. The proposed solution computes an actuation that drives the vehicle into feature rich zones, avoiding localization failure. \protect\subref{fig:slam_works}. Illustrations were made using V-REP by Coppelia Robotics.}
    \label{uncertainty}
\end{figure}

Basically, visual odometry aims at estimating the camera motion by taking the motion of tracked features in two consecutive frames into consideration. Keypoint methods \cite{mur-artal15} extract features from salient image regions, to recover camera pose using epipolar geometry. While direct methods \cite{c3,c5} consider photometric information at high frame rates for a robust pose estimation. For passive VO \cite{scaramuzza11}, interest image regions are initialized and tracked as a consequence of navigation. When visual cues are not available, the vision pipeline fails and state estimation relies on model propagation and inertial data. Within few seconds, state estimation error and uncertainty grows fast in time and vehicle may get lost, as is illustrated in Fig.~\ref{uncertainty}\subref{fig:slam_fails}.  
There are several approaches in the literature, namely active-SLAM \cite{c8,Vidal-Calleja10}, next-best-view \cite{sadat14,c11} and planning under uncertainty \cite{c13,c6} that tackle this problem. The basic idea consists in adding visual information in the motion planning and control loop to minimize state estimation uncertainty. 

Davison and Murray \cite{c8} define the main goal in active feature selection as building a map of features, which helps localizing the robot rather than an end result in itself. In that sense, Vidal-Calleja et al. \cite{Vidal-Calleja10} propose a control law that drives the camera such that expected information gain is maximized. Sadat et al. \cite{sadat14} propose a scoring function that takes into account the expected number of features for a given camera's viewpoint, using a mesh of triangles. Mostegel et al. \cite{c11} propose a set of measurements, including geometric point quality and recognition probability, to analyze the impact of possible camera motions, and avoid localization loss.

Considering a known map (given {\it a priori}), the navigation task can be improved by planning routes that favor texture-rich areas. Achtelik et al. \cite{c13} addresses a Rapid-exploring Random Belief Tree (RRBT) framework that incorporate MAV dynamics and pose uncertainty. In \cite{c6}, pose uncertainty is incorporated in a Rapid-exploring Random Trees (RRT*) framework. Most informative trajectories are selected using Fischer informative matrix. Previous non-mapped or non-static regions have impact on local edges and vertexes affected by new information. 

This paper proposes a low-level strategy for MAVs to improve the locally visual odometry performance. Therefore, high level localization task, such as Simultaneous Localization And Mapping (SLAM) are also benefited. The solution guides the vehicle through a path with high quality image features. In particular, we propose a method based on Artificial Potential Field (APF) where each image feature is associated with a corresponding potential energy. A reference velocity is thus derived that points towards a feature rich region in the current image frame. The total control input velocity is a linear combination of the reference velocity with the one imposed by the motion control algorithm so that the vehicle is driven toward the goal, while avoiding texture-less and non-static features regions, as it is illustrated in Fig. 1(b).

Most active localization solutions addressed in the literature are built on top of SLAM, being relatively complex and map dependent. In contrast, our method does not require a map, but only features selected as inliers in the current frame. Therefore, similarly to the original work on artificial potential fields by Khatib \cite{khatib86}, it is a low level real-time local solution which can be integrated with a high-level method for better performance. In particular, the proposed solution will not experience the same type of errors as in active SLAM solutions and can be integrated in a complementary fashion such that the final solution avoids local minima.

The remainder of this paper is organized as follows: Sec.~II introduces basic notations and definitions. In Sec.~III, the proposed method is presented. Sec.~IV presents experimental results. Finally, Sec.~V addresses final remarks and future work.

\section{Notations and definitions}
\label{sec:notation_definition} 
Consider the 3D body fixed frame $\{B\}$ and the 2D image plane frame $\{I\}$. The origin of $\{B\}$ coincides with the center of gravity of the vehicle and the origin of $\{I\}$ corresponds to the top-left image pixel. Vectors are described in lower case bold and a leading superscript indicates its coordinate frame. The homogeneous coordinate of vector $\mathbf{v}$ is denoted as $\bar{\mathbf{v}}$. When a vector is described in $\{I\}$, the leading superscript is omitted. Matrices are written in upper case and sets in calligraphic letter. The transformation from body to image frame ${T}= K [{R}|\mathbf{t}] \in \mathbb{R}^{3\times 4}$ is known, where $R$ and $\mathbf{t}$ denotes the rotation and translation from $\{B\}$ to $\{C\}$, and K is the intrisic parameter matrix of the camera.
 
A high level positioning controller computes ${^B\mathbf{v}_{g}} = (v_x,v_y,0) \in \mathbb{R}^3$, a velocity that drives the vehicle to the spatial goal. Notice that only planar motion is considered, that is the vehicle keeps a constant height. Also, let $\mathbf{p}=(u,v) \in \mathbb{R}^2$ be the undistorted coordinates of an image feature. In particular, ${\mathbf{p}_i} \in \mathcal{F}$ is the set of features tracked and selected as inliers in the current image frame.

\section{Proposed Method}
\label{sec:proposed_method}
In this section we describe the proposed method for computing the velocity that drives the vehicle towards a spatial goal, while avoiding low feature areas. This is achieved by adding a component to the goal velocity vector ${^B\mathbf{v}_{g}}$ that favors rich regions regarding features. 

\subsection{Features to charge}
Each feature is associated with an attractive or neutral potential energy. Associating similar potential energy to every feature in the image frame is not adequate, since the system could be easily trapped in a local minima or subject to rough changes when new features are extracted. Instead, taking advantage of the fact that the camera provides bearing information, the proposed method considers the orientation of each feature w.r.t. the direction of $\mathbf{v}_g$, i.e. the projection of the goal velocity in the image frame. Thus, as it is expected in a potential field framework, the final goal itself plays a part in the local decision making process.

\begin{figure}[t!]
  \centering
  \includegraphics[width=0.6\linewidth]{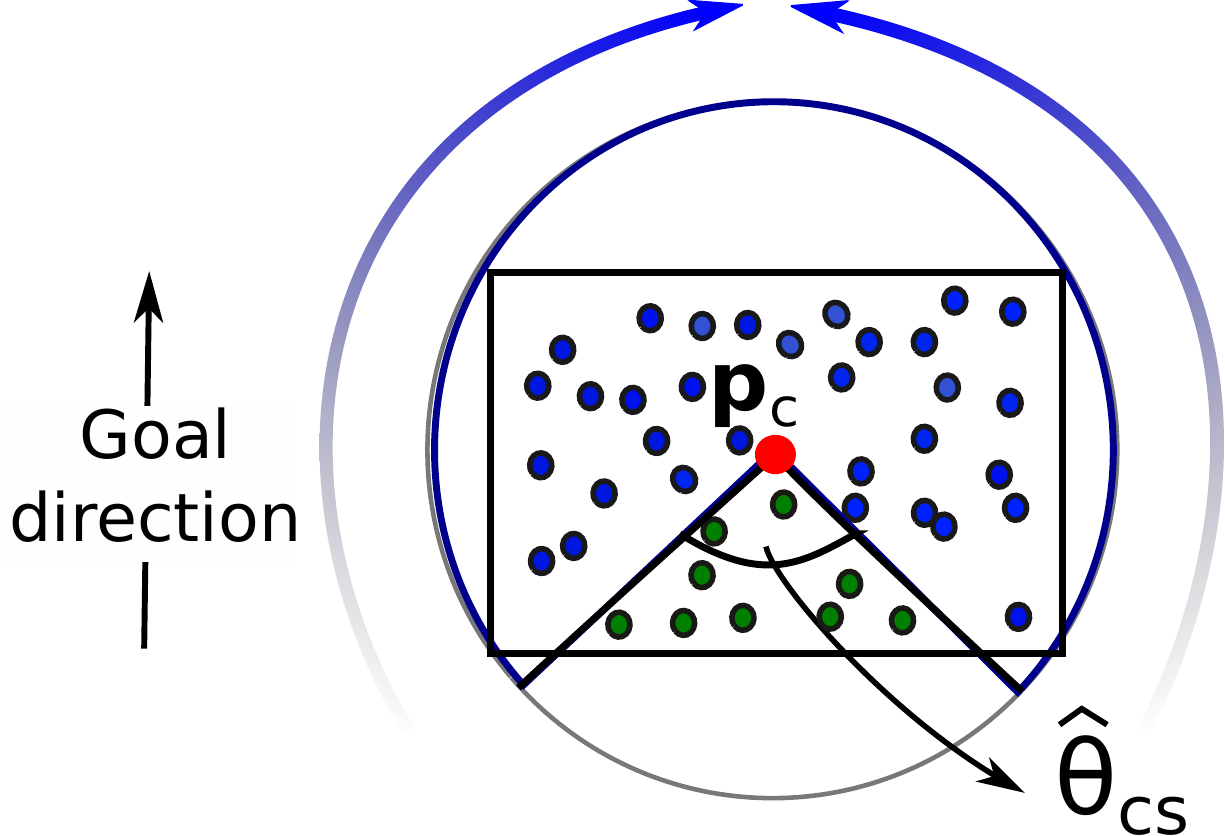}
  \caption{Attractive potential energy increase in the direction of the blue arrow. Charges are neutral within the circle segment defined by $\theta_{cs}$. Attractive and neutral charges are represented by blue and green dots, respectively.}
  \label{fig:energy_assignment}
\end{figure}

Let $\mathbf{p}_c = (u_r, v_r)$ be a point that belongs to the image frame, and consider that the feature based velocity shall be computed at that point. For each feature ${\mathbf{p}_i} \in \mathcal{F}$, compute
\begin{align}
\hat{\mathbf{p}}_i &= (\hat{u}_i, \hat{v}_i) = \mathbf{p}_i -  \mathbf{p}_c \label{eq:p_hat}\\
\theta_i &= \arccos\left(\frac{\langle \hat{\mathbf{p}}_i, {\mathbf{v}_g} \rangle}{\|\hat{\mathbf{p}}_i\|\,\|{\mathbf{v}_g}\|}\right),
\label{eq:theta}
\end{align}
where $\theta_i \in [0,\pi]$ and ${\mathbf{v}}_g \sim {T}\, {^B\bar{\mathbf{v}}_{g}}$. Notice that only the direction of ${^B\mathbf{v}_g}$ is taken into account for computing the feature-based velocity vector. The last coordinate of its homogeneous form must be set to $0$.

Without loss of generality, features in the image plane can be confined within the boundary of a circle centered at $\mathbf{p}_c$, as shown in Fig.~\ref{fig:energy_assignment}. Let the central angle $\hat{\theta}_{cs}$ define a circular segment in the circle, such that the angle $\theta_{cs}$ is defined as 
\begin{equation}
\theta_{cs}=\pi - \hat{\theta}_{cs}/2.
\end{equation}

A charge $Q_i \in \mathcal{Q}$ is represented as the tuple $Q_i=(\hat{\mathbf{p}}_i, q_i, {\theta_{cs}} )$, where $q_i \in [0,1]$ is its corresponding potential energy. The charging policy is defined below:
\begin{equation}
{q}_{_i} =
\arraycolsep=0.4pt\def\arraystretch{1.0}
\left \{ \begin{array}{ll}
1 - \frac{\theta_i}{\theta_{cs}} , & \text{ if } \theta_i \leq \theta_{cs} \\
0 , & \text{ if } \theta_i > \theta_{cs} \\
\end{array} \right. .
\label{eq:q_pos}
\end{equation}

Based on the current frame information, the charging policy considered associates high attractive potential energy to features localized in the goal direction. Features localized away from the goal direction, on the region defined by the circle segment, have a neutral charge.

\subsection{Vector Field}
Each charge $Q_i \in \mathcal{Q}$ exerts a force  $\mathbf{f}_{i}$ at $\mathbf{p}_c$, given by
\begin{equation}
\mathbf{f}_{i} =
\arraycolsep=0.4pt\def\arraystretch{1.0}
\left \{ \begin{array}{ll}
(0,\ 0)^T , & \text{ if } d_i < r \\
\frac{(d_i-r)}{s}\ q_i\Big(\cos(\phi_i),\ \sin(\phi_i) \Big) ^T , & \text{ if } r \leq d_i \leq s + r \\
q_i\Big( \cos(\phi_i),\ \sin(\phi_i) \Big) ^T , & \text{ if } d > s + r
\end{array} \right. ,
\end{equation}
where $r$ is the distance in pixels that a charge must be, from the evaluated point, to exerted any force on it; $s$ is the spread in pixels of the potential field, and $d_i$ and $\phi_i$ are computed as follow
\begin{align}
    d_i &=\|\hat{\mathbf{p}}_i\| \label{eq:d} \\
    \phi_i &= \text{atan2}(\hat{u}_i, \hat{v}_i).
\end{align}

\begin{figure*}[t!]
  \includegraphics[width=0.95\textwidth,height=8cm,keepaspectratio]{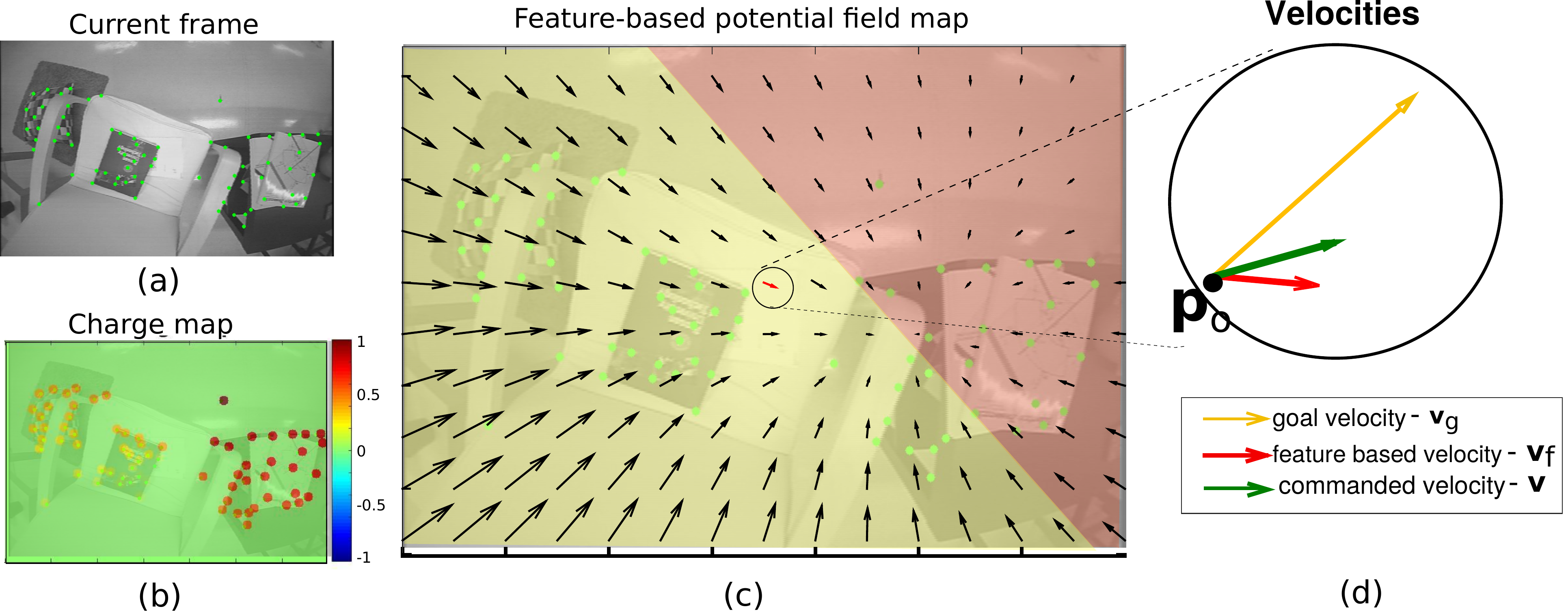}
  \caption{Case study for the feature based potential field. Current image frame and inlier features (a). Considering $\mathbf{p}_c$ as the optical center, the charge map is built and represented as a heat-map. (b). The potential field map shows the action derived when evaluating at different $\mathbf{p}_c$ (c). Goal-friendly region is shown in yellow and feature-friendly in red (more detail about these regions is shown in Sec.~\ref{sec:our_method_discussion}). (d) shows commanded actuation as the combination of goal and feature velocity. ($\mathbf{p}_c=\mathbf{p}_o$, $\lambda = 0.1, \theta_{CS} = 10^{o.}, d=50$ pixel, $s=150$ pixel).}
  \label{fig:showcase}
\end{figure*}

The total force $\mathbf{f}$ on the point $\mathbf{p}_c$ can be computed as
\begin{equation}
\mathbf{f} = \displaystyle\sum_{i}^{} \mathbf{f}_{i},
\end{equation}
which can be normalized and transformed into a feature based velocity command ${\bar{\mathbf{v}}_{f}}$. Its direction in homogeneous coordinates is given as
\begin{equation}
{\bar{\mathbf{v}}_{f}} = \frac{1}{\|\mathbf{f}\|}
\begin{bmatrix}   
\mathbf{f} \\
0
\end{bmatrix}.
\end{equation}
%ensuring that $\|\bar{\mathbf{v}}_{f}\| \leq 1$.

Finally, the proposed command action takes the form
\begin{equation}
{\bar{\mathbf{v}}} = \lambda \bar{\mathbf{v}}_{g} + (1-\lambda)\bar{\mathbf{v}}_{f},
\label{eq:proposed_command} 
\end{equation}
where $\lambda$ is a weight factor and ${\bar{\mathbf{v}}_{g}}$ is a normalized velocity. The velocity ${\bar{\mathbf{v}}}$ can be transformed from the image frame to the body frame and scaled accordingly.

\subsection{Discussion}
\label{sec:our_method_discussion}

Artificial potential fields frameworks usually take into consideration the robot position when computing forces - obstacles exert repulsive force and the goal an attractive force. In the proposed framework, features can be either attractive or neutral accordingly to their position w.r.t. the point $\mathbf{p}_c$ being considered and the goal direction. As for now, only attractive and neutral charges are admitted, in the future, the effect of associating repulsive charge to features classified as outlier will be analyzed.
 
Figure~\ref{fig:showcase} illustrates a case study. In particular, Fig.~\ref{fig:showcase}(a) shows a frame and extracted features classified as inliers. The goal velocity ${\mathbf{v}_g}$ is directed towards the top-right pixel in the image frame - a poor zone regarding the number of features. Fig.~\ref{fig:showcase}(b) depicts the potential energy associate to each feature when evaluating the action induced at the central pixel of the camera, $\mathbf{p}_o$. The potential field map (see Fig.~\ref{fig:showcase}(c)) shows the corresponding field for different values of $\mathbf{p}_c$. For each point in the map, \eqref{eq:p_hat}--\eqref{eq:proposed_command} must be computed. However, for visualization, $\mathbf{v}_{f}$ is not normalized. The map can be classified in a \textit{goal-friendly} and \textit{feature-friendly} actuation zone. Both regions are shown in the potential field map in yellow (goal-friendly) and red (feature-friendly) background. Suppose $\lambda = 0$, then according to \eqref{eq:proposed_command} the vehicle follows $\mathbf{v}_{f}$. If $\mathbf{p}_c$ is within a feature-friendly region, the vehicle favors more the features than the goal. As a matter of fact, the vehicle will move away from the goal. On the contrary, if $\mathbf{p}_c$ is within a goal-friendly region, the vehicle will move towards the goal. The radius of the charge $r$ determines whether features close to the point $\mathbf{p}_c$ affect the solution or not. Meanwhile, the spread $s$ determines the strength of each charge. The larger the spread is, the more influence charges away from the point evaluated will have on the feature driven action. Thus, it limits the prediction horizon based on the local frame, that is, the belief that a feature on an edge indicates there will be more features on that direction.

\section{Experimental Results}
This section describes the experimental setup, as well as the results observed for multiple flight tests.

\subsection{Experimental Setup}
\begin{figure}[t]
  \centering
  \includegraphics[width=0.7\linewidth]{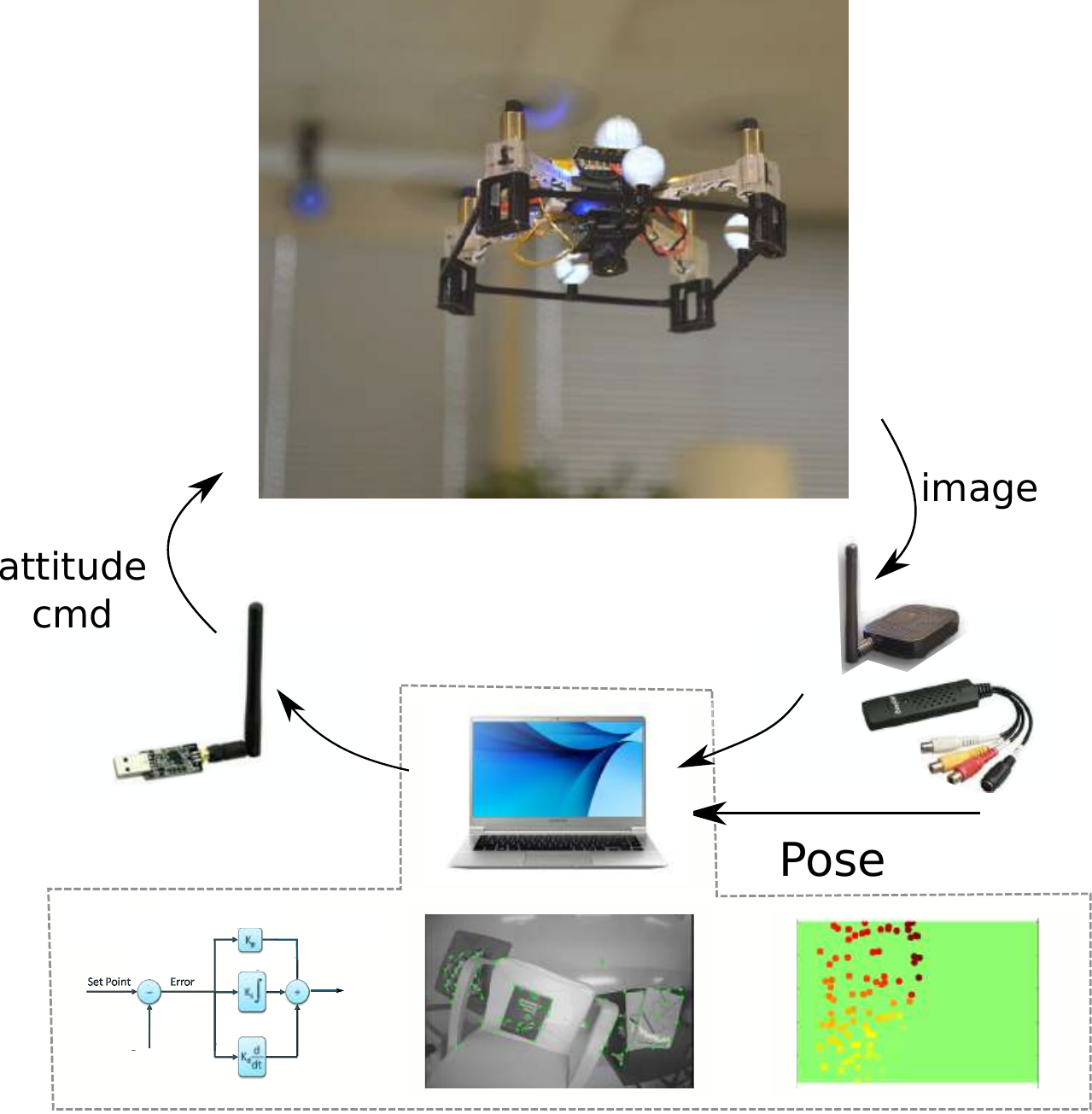}
  \caption{Experimental setup. Clockwise, starting from the top: the crazyflie equipped with the mini transmitter camera module FX798T, image receiver, and capture card; a notebook that runs the code; and an antenna for the communication with the vehicle.}
  \label{fig:experimental_setup}
\end{figure}

Fig.~\ref{fig:experimental_setup} shows the main components employed in experimental validation. The algorithm was tested in the mini-quadrotor Crazyflie 2.0, manufactured by BitCraze. We attached to the vehicle a $4.7\,g$ mini transmitter camera module FX798T equipped with $120^o$ field of view lens - the camera faces downward. This module broadcasts images using an embedded $5.8\,GHz$ transmitter. On the other side, we use a receiver that is connected to a computer through a video capture card. The frame-rate is $30\,Hz$ for a $720\times 480$ image resolution. The proposed algorithms runs on the off-board computer at frame-rate speed. The flying arena is equipped with an Optitrack, motion capture system manufactured by the company NaturalPoint. The system records the motion of retro-reflective markers using the optical-passive technique. Infrared cameras capture markers' position allowing a pose estimation with millimetric precision. 

Images and commands are published and received within the ROS (Robot Operating System) environment. Collective thrust and attitude commands are sent to the vehicle using the package developed in \cite{hoenig15}. A customized PID controller assures the vehicle follows the desired action. The motion capture system provides feedback for the control loop, while the feature-based potential field works online as the decision making process, sending desired velocity commands to the controller. A first order low pass filter with cutoff frequency at $20\ Hz$ provides smooth control reference for the feature driven action.

When a new image frame is received, Shi-Tomasi features \cite{shi94} are extracted. In addition to a minimum quality threshold, only $100$ features with the highest response are selected. Then, features are tracked across two consecutive frames using Lucas-Kanade Tracker (LKT) \cite{lucas81}. Within a RANSAC \cite{fischler81} framework, the 8-point algorithm classifies features as inliers or outliers. The solution is robust under the presence of few false inliers, such that a matching step is unnecessary.

\begin{figure}[t]
  \centering
  \includegraphics[width=0.7\linewidth]{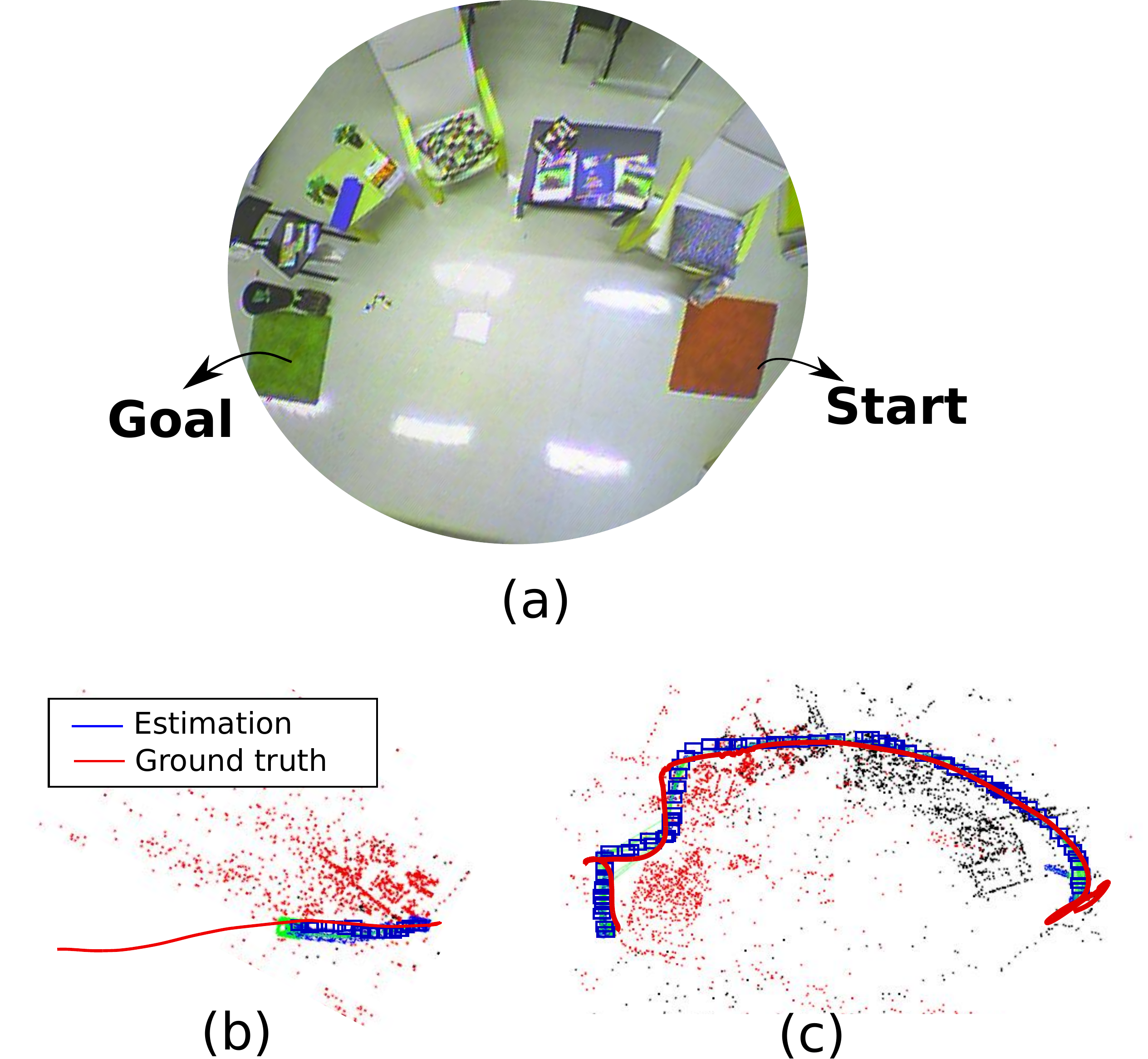}
  \caption{Scenario for evaluation of proposed algorithm (a). Localization fails in the absence of features (b). Feature based velocity vector drives the vehicle to the goal through a region rich in features (c).}
  \label{fig:proof_of_concept}
\end{figure}

\subsection{Evaluation}
\begin{figure}[t]
  \centering  \includegraphics[width=0.75\linewidth]{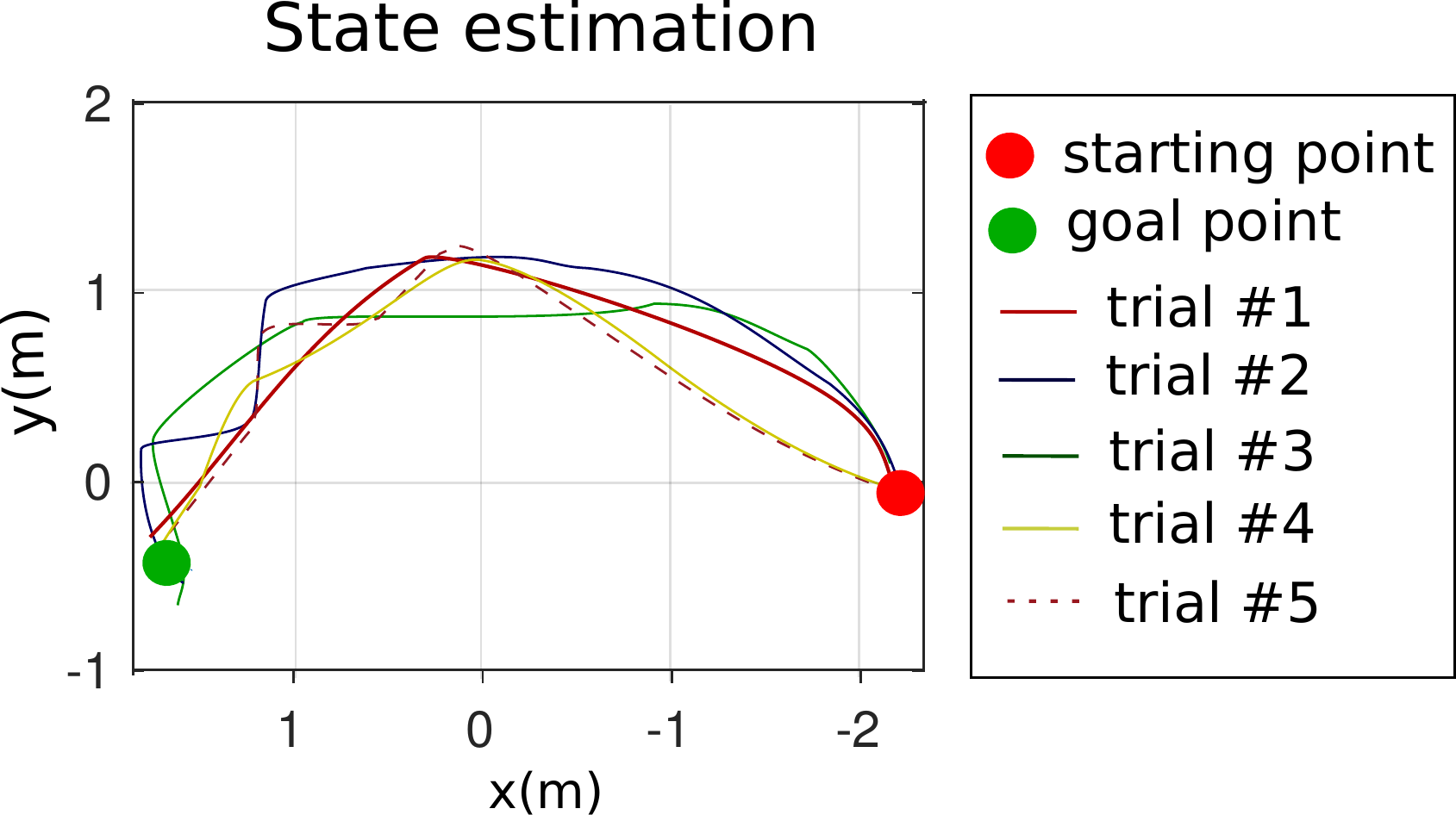}
    \caption{Multiple trial for $\lambda=0.45$, $\hat{\theta}_{cs}=30^o$ and $\mathbf{p}_c = \mathbf{p}_o$.}
    \label{fig:multiple_trials}
\end{figure}

Multiple flights were performed in a $4\,m\times3\,m$ living room alike arena, as shown in Fig.~\ref{fig:proof_of_concept}(a). The robot was autonomously controlled to fly from the starting position to the goal position as indicated by the two carpets. As seen from the image, the straight path that links both carpets is a regular floor that does not contain good visual cues. The localization task for a trial is said to $succeed$ if ORB-SLAM \cite{mur-artal15} manages to keep track of the pose of the vehicle. ORB-SLAM run offline, just for evaluation purpose. Multiple trials showed that taking the straight path towards the goal always results in localization failure due to the lack of visual features. Fig.~\ref{fig:proof_of_concept}(b) shows one of the trials where the localization estimation fails when a straight path is taken. Alternatively, using the proposed active method, the robot was always capable of reaching the goal while maintaining localization. Fig.~\ref{fig:proof_of_concept}(c) shows the path the robot takes, as well as the estimated state throughout the time in one of the experiments. Fig.~\ref{fig:multiple_trials} shows different trials with the proposed method where the robot reaches the goal while keeping track of its location. 

\section{Conclusions}
This paper proposes a low-level solution for the active visual navigation problem. Features tracked across consecutive frames are associated with an attractive or neutral potential energy. The intensity of each charge is a function of the goal direction, which is given by a motion-control algorithm. The desired vehicle velocity combines a component that takes visual cues into consideration and a component related to the spatial goal. Experimental results using a micro aerial vehicle, equipped with a downward looking camera, showed that the method can effectively drive the vehicle towards the goal while avoiding no or poor featured regions. The proposed active solution does not rely on a map and hence can be integrated within a SLAM framework to improve the accuracy and robustness of localization and mapping.  
Points that are currently being investigated include the analysis of the effect of dynamic tuning of parameters based on mission \& environment characteristics, associating repulsive energy to features classified as outliers, and the reasoning about computing the feature based velocity vector in other points than the camera's optical center.

\section*{Acknowledgement}
R. Rogrigues, M. Basiri, and P. Miraldo were partially supported by Funda\c{c}\~{a}o para a Ci\^{e}ncia e Tecnologia (FCT) project UID/EEA/50009/2013, and by FCT grant SFRH/BPD/111495/2015. P. Aguiar was partially supported by project POCI-01-0145-FEDER-006933/SYSTEC funded by FEDER funds through COMPETE2020 - Programa Operacional Competitividade e Internacionaliza\c{c}ç\~{a}o (POCI) - and by national funds through FCT.

\end{document}